\documentclass{article}

\usepackage[preprint]{neurips_2026}

\usepackage[utf8]{inputenc} % allow utf-8 input
\usepackage[T1]{fontenc}    % use 8-bit T1 fonts
\usepackage{hyperref}       % hyperlinks
\usepackage{url}            % simple URL typesetting
\usepackage{booktabs}       % professional-quality tables
\usepackage{amsfonts}       % blackboard math symbols
\usepackage{nicefrac}       % compact symbols for 1/2, etc.
\usepackage{microtype}      % microtypography
\usepackage{graphicx}
\usepackage{amsmath}
\usepackage{amssymb}
\usepackage{booktabs}
\usepackage{multirow}
\usepackage[table]{xcolor}
\title{AnchorDiff: Topology-Aware Masked Diffusion with Confidence-based Rewriting for Radiology Report Generation}
\author{%
  \begin{tabular}{c}
  Shiying Yu \quad Jielei Wang \quad Guoming Lu \\
  University of Electronic Science and Technology of China
  \end{tabular}
  }

\begin{document}

\maketitle

\begin{abstract}
  Radiology report generation (RRG) aims to automatically produce clinically accurate textual reports from medical images. Existing
  methods predominantly rely on autoregressive (AR) language models, whose causal dependency structure restricts generation to a
  unidirectional left-to-right process. This paradigm can induce sequence bias, where models tend to follow stereotypical token
  orders and high-frequency report templates rather than fully grounding generation in image-specific evidence. In this paper, we
  propose \textbf{AnchorDiff}, the first masked-diffusion framework for RRG that integrates knowledge-graph-derived clinical anchors
  into diffusion language modeling. By leveraging bidirectional context and iterative refinement, AnchorDiff mitigates the
  limitations of fixed-order autoregressive decoding. Specifically, we introduce a topology-aware training strategy that uses
  RadGraph-derived entity hierarchies to assign clinically important tokens differentiated masking protection and loss weights. We
  further design an inference-time rewriting strategy that detects unstable committed tokens through perturbation-based testing and
  selectively revises them during denoising. Extensive experiments on the MIMIC-CXR and MIMIC-RG4 benchmarks demonstrate that
  AnchorDiff achieves state-of-the-art (SOTA) performance, showing the effectiveness of clinically anchored masked diffusion for radiology
  report generation.
\end{abstract}

\section{Introduction}

The proliferation of advanced medical imaging modalities has generated an unprecedented volume of diagnostic data, vastly outpacing the growth of the professional radiologist workforce. 
Consequently, radiologists are routinely burdened with interpreting hundreds of scans daily and drafting extensive, highly repetitive reports. 
This escalating workload inevitably leads to diagnostic fatigue, elevating the risk of overlooking subtle or rare pathological findings. 
To address this clinical need, automated radiology report generation (RRG) has attracted growing research attention. 
Given one or more medical images as input, the goal of RRG is to produce a coherent, clinically accurate free-text report that describes both normal anatomy and any abnormal findings.

Early encoder--decoder RRG methods mainly improve image--text alignment and report coherence through memory-augmented cross-modal interaction, disease-tag or region-level visual grounding, knowledge-guided generation, and prototype or expert-token-based representation learning \citep{b1,b2,b45,b12}, yet remain limited by their relatively small model capacity. More recently, driven by the strong emergence of high-quality Vision-Language Models (VLMs), a series of cutting-edge methods, including LLM-CXR \citep{b3}, MAIRA-1 \citep{b4}, and LLM-RG4 \citep{b5} have achieved substantial breakthroughs by leveraging the rich medical knowledge 
\begin{figure}
  \centering
  \includegraphics[width=1.0\linewidth]{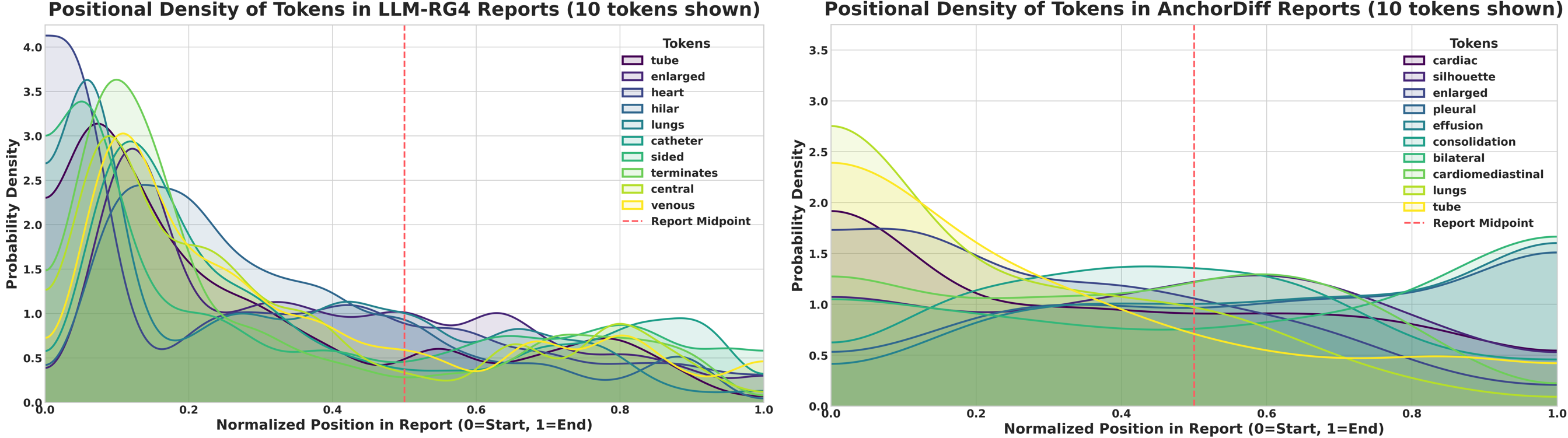}\href{}{}
  \caption{Based on the word frequency distribution within the reports generated by the autoregressive model, it can be observed that pathological terms are predominantly concentrated in the first half.}
  \label{fig:1}
  \vspace{-15pt}
\end{figure}
embedded in pretrained large language models (LLMs). Despite their architectural diversity, these methods universally adopt the autoregressive (AR) paradigm, in which the generation of each token is strictly conditioned on its left context. We argue that this unidirectional dependency structure introduces a systematic failure mode in the RRG setting, which manifests as typical sequence bias: because a large proportion of reports in standard benchmarks such as MIMIC-CXR \citep{b6} describe normal or frequently recurring findings, the AR conditional distribution $P(w_t \mid w_{<t})$ can be biased toward stereotypical report continuations. This tendency is further evidenced by the model's pronounced concentration of pathological terms within the earlier segments of generated reports, as illustrated in Figure~\ref{fig:1}. More specifically, this bias gives rise to three interrelated model deficiencies: (i) \textbf{statistical shortcutting}, where the model exhibits a preference for generating tokens in a fixed order to evade loss penalties rather than attending to image-specific visual evidence; (ii) \textbf{causal chain lock-in}, where an early erroneous token irreversibly propagates through the remainder of the sequence and (iii) \textbf{global coherence deficiency}, where the model cannot jointly reason about spatially distributed pathological findings like bilateral lung lesions, because each position only accesses preceding tokens.

Motivated by these observations, in this paper we depart from the AR paradigm and instead adopt masked diffusion language models (MDLMs) \citep{b8} as the generation backbone for RRG. We posit that the iterative mask-and-predict mechanism of MDLMs is inherently aligned with how radiologists compose reports in practice: rather than writing strictly from left to right, experienced radiologists first identify major anatomical landmarks and salient pathological findings across the entire image, then progressively refine the descriptive details, a cognitive process that is fundamentally non-sequential. Specifically, MDLMs offer three structural advantages that architecturally bypass the unidirectional dependency flaws of the aforementioned AR models. First, they provide bidirectional context: every token is predicted conditioned on the full sequence in both directions, enabling the model to leverage image-grounded evidence from all positions rather than relying on left-only statistical shortcuts. Second, they enable confidence-driven progressive revelation: tokens with higher prediction confidence are unmasked earlier in the denoising process, naturally prioritizing definitive findings over uncertain details and thus preventing any single premature decision from locking the remainder of the sequence. Finally, MDLMs exhibit inherent anchor affinity: the iterative refinement nature of diffusion allows differential treatment of tokens across denoising steps, making it possible for clinically important tokens to serve as stable structural anchors around which the rest of the report is coherently organized.

Building on these insights, we propose AnchorDiff, a masked-diffusion framework for radiology report generation that introduces two complementary mechanisms targeting the training and inference stages, respectively. On the training side, we design AnchorTree, a topology-aware strategy that leverages the relational structure between entities extracted by RadGraph \citep{b9} to construct a three-level clinical hierarchy over report tokens anatomy as root anchors, core pathological observations as intermediate nodes, and qualitative modifiers as leaves. This hierarchy is injected into the diffusion process through differentiated masking protection and anchor-weighted loss supervision, teaching the model to treat clinically critical tokens as structural scaffolding around which the remaining content is denoised. On the inference side, we introduce a lightweight self-correction mechanism tailored for DLMs. At intermediate denoising steps, Clinically-Aware Perturbation-based Token Rewriting (CAPTR) employs controlled perturbation to identify prematurely committed tokens: if a token fails to maintain its initial prediction upon perturbation, exhibiting contextual instability, it is resampled. 

\begin{itemize}
    \item We propose \textbf{AnchorDiff}, to the best of our knowledge the first framework that adopts masked diffusion language models for radiology report generation, fundamentally addressing the sequence bias inherent in autoregressive approaches.
    
    \item We design \textbf{AnchorTree}, a topology-aware training mechanism that converts RadGraph entity hierarchies into token-level masking protection and loss weighting signals, enabling the diffusion process to be guided by clinically meaningful anchor structures.
    
    \item We introduce \textbf{Clinically-Aware Perturbation-based Token Rewriting mechanism}, a training-free inference-time self-correction strategy that identifies and revises prematurely committed tokens through perturbation-based instability testing, enhancing report coherence at minimal computational cost.
    
    \item Extensive experiments on the MIMIC-CXR benchmark demonstrate that AnchorDiff achieves state-of-the-art (SOTA) performance on both natural language generation and clinical efficacy metrics.
\end{itemize}

\section{Related Work}
\textbf{Radiology Report Generation.} Research on automated radiology report generation has evolved along two main axes: architectural design for image-text alignment and the incorporation of domain knowledge. Along the first axis, R2Gen\citep{b1} introduces a relational memory to capture inter-sentence dependencies, while R2GenCMN\citep{b2} extends this with a cross-modal memory network that maintains shared visual-textual representations across decoding steps. METransformer \citep{b12} further improves alignment through entity-level multi-expert modules that decompose the generation task by anatomical region. With the emergence of vision-language models, recent methods have shifted toward leveraging pretrained LLMs as the text generation backbone. LLM-CXR \citep{b3} directly fine-tunes an instruction-tuned LLM on radiology image-text pairs via visual token injection. MAIRA \citep{b4} pairs a radiology-specific vision encoder with a GPT-class decoder and introduces grounded report generation with bounding-box supervision. LLM-RG4\citep{b5} proposes an adaptive token fusion mechanism that accommodates four clinical input scenarios single or multi-view, with or without prior reports — achieving strong performance across all settings on MIMIC-CXR \citep{b6}. Along the second axis, several works integrate clinical knowledge through medical knowledge graphs \citep{b13} or RadGraph-based reward signals for reinforcement learning \citep{b14}. Notably, despite their diverse designs, all existing methods share the autoregressive generation paradigm, and domain knowledge has been employed exclusively at the output level  for evaluation metrics or auxiliary rewards rather than being used to directly structure the training process of the generative model itself.

 \textbf{Diffusion Models for Text Generation}
  Diffusion models have achieved remarkable success in continuous domains such as image synthesis \citep{b15, b16}, and
  a growing body of work has explored their adaptation to discrete text generation. Continuous-space approaches, such as
  Diffusion-LM \citep{b17} and DiffuSeq \citep{b18}, embed tokens into continuous space, apply Gaussian noise, and then
  denoise to generate text. While pioneering, these methods suffer from rounding errors during the continuous-to-
  discrete conversion and scale poorly to long sequences. Discrete-space formulations circumvent this issue by operating
  directly on token identities. D3PM \citep{b19} establishes the theoretical foundation for discrete diffusion via
  structured transition matrices, and MDLM \citep{b20} simplifies the framework to a masked-predict objective akin to
  BERT but with generative capability. Most recently, LLaDA \citep{b21} scales this paradigm to 8 billion parameters,
  demonstrating that masked diffusion language models can match the performance of comparably sized autoregressive
  models (e.g., LLaMA 3 8B) on general-purpose benchmarks spanning reasoning, code, and instruction following.
  Recent work has also explored inference-time revision for diffusion language models. CoRE~\citep{b10}
  proposes context-robust remasking to identify unstable committed tokens, but it is designed for general text
  generation
  and does not incorporate clinical entity structure.
  In the medical domain, ECHO \citep{b26} has recently explored diffusion-based chest X-ray report generation from the
  perspective of runtime efficiency, showing the feasibility of applying DLMs to RRG. However, its primary focus is
  efficient decoding, whereas our work targets a different and complementary problem: enforcing clinically structured
  generation and improving the content quality of reports. Specifically, existing diffusion-based RRG methods still lack
  explicit mechanisms for organizing anatomical anchors, pathological findings, and descriptive modifiers as structured
  constraints during denoising. Our work bridges this gap by adapting LLaDA \citep{b21} to the RRG task and further introducing
  topology-aware training (AnchorTree) and Clinically-Aware Perturbation-based Token Rewriting mechanism (CAPTR) to
  address the structural and clinical fidelity demands of radiology report generation.
\label{gen_inst}

\begin{figure}
  \centering
  \includegraphics[width=1.0\linewidth]{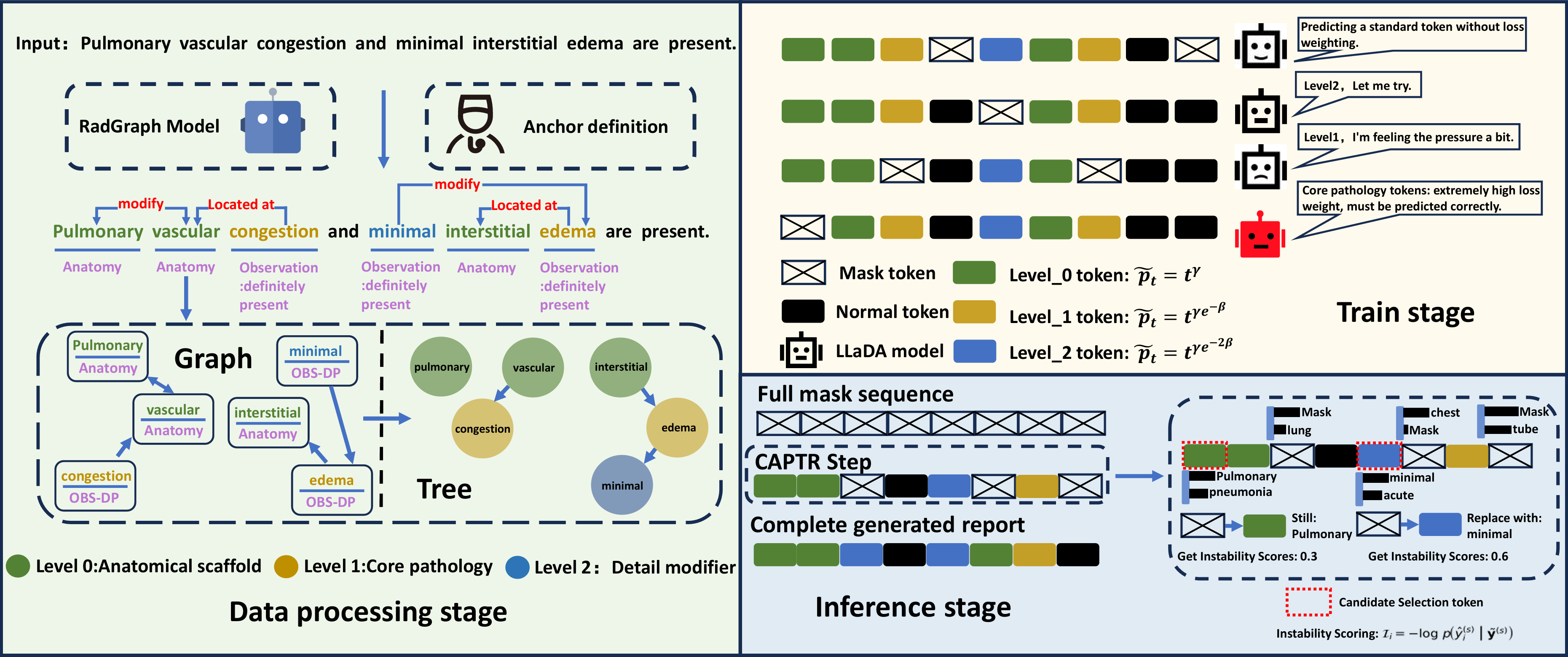}\href{}{}
  \caption{\textbf{Overview of AnchorDiff.} Clinical entities extracted by RadGraph are organized into a hierarchical anchor tree and assigned level-aware masking weights for LLaDA training. During inference, CAPTR progressively refines unstable tokens to generate clinically consistent radiology reports.}
  \label{fig:framework}
\end{figure}

\section{Method}
\label{headings}
\subsection{Problem Formulation and Framework Overview}

\textbf{Problem Formulation.} Formally, the task of radiology report generation takes as input a frontal chest X-ray image
$x_f$, alongside an optional lateral view $x_l$ and a prior report $r_p$. The objective is to generate a target
textual sequence $\mathbf{y} = \{y_1, \ldots, y_T\}$ that accurately describes the clinical findings. As illustrated
in Figure\ref{fig:framework}, our approach, AnchorDiff, departs from standard autoregressive paradigms by adopting
LLaDA \citep{b21}, a masked diffusion language model, as the generative backbone. To address the unique demands of
clinical text, AnchorDiff introduces two key mechanisms: \textit{AnchorTree}, which restructures the masking
schedule and loss function according to clinically grounded token hierarchies, and \textit{CAPTR}, which identifies
and revises unreliable predictions during inference.

\textbf{Multi-Modal Conditioning and Training.} Before the diffusion process, heterogeneous inputs must be encoded
into a unified space. We extract visual features from $x_f$ (and $x_l$ if available) using a frozen RAD-DINO
\citep{b22}, and encode $r_p$ via a frozen CXR-BERT\citep{b23}. These modality-specific features are
projected into LLaDA's embedding space via cross-attention modules and fused into a unified representation
$\mathbf{V} \in \mathbb{R}^{N_q \times d}$, where $N_q$ denotes the number of learnable query tokens and $d$ is the
hidden dimension of LLaDA. Crucially, $\mathbf{V}$ serves as a continuous condition that steers LLaDA's discrete
mask-and-predict diffusion steps. The entire framework is trained in a two-stage curriculum: Stage1 aligns the
visual projectors with the frozen LLaDA on single-view data; Stage 2 jointly fine-tunes all components via
LoRA \citep{b11} across all multi-view and historical-report scenarios.

\subsection{AnchorTree: Topology-Aware Diffusion Training} \label{sec:anchortree}

\textbf{Masked Diffusion Background.} LLaDA formulates text generation as a discrete denoising process. In the forward process, a clean token sequence $\mathbf{y} = \{y_1, \ldots, y_T\}$ is progressively corrupted by independently masking each token with probability $t \in [0, 1]$:
\begin{equation}
    q\!\left(z_t^i \mid y_i\right) =
    \begin{cases}
        \text{[MASK]} & \text{with probability } t \\
        y_i           & \text{with probability } 1-t
    \end{cases}
\end{equation}
The model $p_\theta$ is trained to reconstruct all masked tokens from the corrupted sequence $\mathbf{z}_t$, minimizing the standard masked language modeling objective:
\begin{equation}
    \mathcal{L}_{\text{base}} = \mathbb{E}_{t,\, \mathbf{z}_t} \left[ -\sum_{i \in \mathcal{M}_t} \log p_\theta(y_i \mid \mathbf{z}_t) \right]
\end{equation}
where $\mathcal{M}_t = \{i : z_t^i = \text{[MASK]}\}$. During inference, generation proceeds over $S$ steps from a fully masked sequence. At each step, the model predicts all remaining masked positions simultaneously and commits a subset of most confident tokens, enabling non-autoregressive parallel generation. The above uniform masking treats all tokens identically, ignoring the intrinsic clinical structure of radiology reports. AnchorTree addresses this by restructuring both the masking schedule and the training objective based on a clinically grounded token hierarchy.

\textbf{RadGraph Anchor Hierarchy.} We parse each training report using RadGraph to extract a directed entity-relation graph, assigning each token a discrete anchor level $\ell \in \{0, 1, 2, -1\}$. Tokens of anatomical entities (e.g., \textit{heart}, \textit{lung}) are designated Level~0; tokens of non-modifier pathological observations (e.g., \textit{effusion}, \textit{pneumothorax}) are Level~1; observation tokens that issue \textit{modify} relations (e.g., \textit{mild}, \textit{bilateral}) are Level~2; and all remaining tokens are Level~$-1$ (non-anchor). This hierarchy reflects the clinical dependency structure: anatomical anchors provide the spatial scaffold upon which observations and modifiers are grounded.

\textbf{Topology-Aware Masking.} We define a level-decay factor $\eta(\ell) = \exp(-\beta \ell)$ and set the effective masking probability for token $i$ as:
\begin{equation}
    \tilde{p}_t(i) = t^{\phi(i)}, \quad
    \phi(i) = \begin{cases}
        \gamma \cdot \eta(\ell_i) & \text{if } \ell_i \geq 0 \\
        1                         & \text{otherwise}
    \end{cases}
\end{equation}
where $\beta = 1.0$ governs the decay rate across levels and $\gamma = 1.5$ controls protection strength. Since $t \in (0,1)$, exponent $\phi > 1$ yields $\tilde{p}_t < t$, which explicitly protects Level~0 anatomy tokens from masking; conversely, $\phi < 1$ at deeper levels increases the effective masking rate. This differential masking compels the model to reconstruct uncertain pathological observations and modifiers conditioned on the visible anatomical scaffolding.

\textbf{Hierarchical Loss Weighting.} To ensure the structural integrity of the generation, the cross-entropy loss $\mathcal{L}_i = -\log p_\theta(y_i \mid \mathbf{z}_t)$ at each masked position is scaled by:
\begin{equation}
    w(i) = \begin{cases}
        1 + \lambda \cdot \eta(\ell_i) & \text{if } \ell_i \geq 0 \\
        1                              & \text{otherwise}
    \end{cases}
\end{equation}
The overall training objective becomes:
\begin{equation}
    \mathcal{L} = \mathbb{E}_{t,\, \mathbf{z}_t} \left[ \sum_{i \in \mathcal{M}_t} w(i) \cdot \mathcal{L}_i \right]
\end{equation}
with $\lambda = 1.1$ (The ablation study regarding these three parameters($\beta$,$\gamma$) is presented in Section~\ref{sec:Ablation}). Notably, although Level~0 anchors are masked less frequently due to Eq.~3, they receive the highest loss multiplier ($1 + \lambda = 2.1$) when they do fall into $\mathcal{M}_t$. This deliberate design imposes a substantially higher penalty on errors involving foundational structural tokens, ensuring the model assigns prioritized attention to clinically critical anchors throughout training.

\subsection{Clinically-Aware Perturbation-based Token Rewriting} \label{sec:core}

Even with AnchorTree training, the iterative nature of masked diffusion inference can leave certain tokens committed prematurely to predictions that ultimately prove suboptimal given the full context. Once a token is unmasked in an early step, it becomes part of the fixed context for subsequent denoising steps—effectively propagating errors if the initial commitment was unreliable. To mitigate this cascading effect, we integrate CAPTR, a training-free inference-time procedure that periodically identifies and revises low-confidence token commitments.

\textbf{Triggering Condition.} Let $\hat{\mathbf{y}}^{(s)}$ denote the partially generated sequence at inference step $s$ (out of $S = 80$ total steps), and let $\mathcal{U}^{(s)}$ denote the set of already-unmasked token positions. CAPTR is activated every $E = 8$ steps within the progress window $s/S \in [0.25, 0.75]$, triggering five times in total. This window is strategically chosen to avoid the early phase—where revealed tokens are too sparse to provide meaningful evaluation context—and the final phase, where revisions near completion carry disproportionate stability risks.

\textbf{Candidate Selection.} At each trigger, we assess the reliability of every revealed token $i \in \mathcal{U}^{(s)}$ via its confidence margin:
\begin{equation}
    \Delta_i = P^{(1)}_i - P^{(2)}_i
\end{equation}
where $P^{(1)}_i$ and $P^{(2)}_i$ are the top-two predicted probabilities for position $i$ under the current forward pass. A smaller $\Delta_i$ indicates higher uncertainty in the committed token. We select the $M = 3$ positions with the smallest margin as candidates for further instability analysis.

\textbf{Perturbation and Instability Scoring.} For the selected candidate set $\mathcal{C}^{(s)}$, we construct a perturbed sequence $\tilde{\mathbf{y}}^{(s)}$ by re-masking exactly those positions, then run one additional forward pass to compute the \textit{instability score}:
\begin{equation}
    \mathcal{I}_i = -\log\, p\!\left(\hat{y}_i^{(s)} \;\middle|\; \tilde{\mathbf{y}}^{(s)}\right), \quad i \in \mathcal{C}^{(s)}
\end{equation}
A high $\mathcal{I}_i$ indicates that the original token $\hat{y}_i^{(s)}$ is poorly supported by its surrounding context once removed, revealing a lack of contextual resilience.

\textbf{Selective Revision.} We identify the $K = 1$ candidate with the highest instability score for potential revision. If the model's top-1 prediction $\hat{y}_i^{\prime}$ under the perturbed context differs from the current token and satisfies $\tilde{P}^{(1)}_i \geq \tau = 0.3$, we replace $\hat{y}_i^{(s)}$ with $\hat{y}_i^{\prime}$. This conservative dual criterion—requiring both a prediction shift and sufficient confidence—ensures that revisions are driven by meaningful contextual evidence rather than random noise. The entire CAPTR procedure requires only 5 additional forward passes across 80 inference steps, incurring a minimal computational overhead of approximately 6.25\%.

\section{Experimental Setup}
\subsection{Datasets and Metrics}
\textbf{Datasets.} We evaluate AnchorDiff on two benchmark settings. 
First, we use \textbf{MIMIC-CXR} \citep{b6}, the largest publicly available chest X-ray dataset with over 227,000 radiograph-report pairs, under the conventional radiograph-only setting to assess the basic report generation capability of our model. 
Second, we adopt \textbf{MIMIC-RG4} \citep{b5}, a multi-context benchmark constructed based on MIMIC-CXR, to evaluate AnchorDiff under more realistic clinical scenarios. 
MIMIC-RG4 extends the standard evaluation protocol by considering both the number of radiographic views and the availability of historical patient records, resulting in four practical settings: \textbf{SN} (single-view, no prior), \textbf{SW} (single-view, with prior), \textbf{MN} (multi-view, no prior), and \textbf{MW} (multi-view, with prior). 
This setting enables a more comprehensive assessment of model robustness across dynamic clinical contexts.

We evaluate model performance across two dimensions. For natural language generation (NLG), we report BLEU-1/2/3/4 (B@n) \citep{b28}, METEOR (MTR) \citep{b29}, and ROUGE-L (R-L) \citep{b30}. For clinical efficacy (CE), we adopt CheXbert~\citep{b24} to extract diagnosis labels and compute micro-averaged Precision (P), Recall (R), and F$_1$ score.

\subsection{Baselines}                                                                                                
  For the standard MIMIC-CXR setting, we compare AnchorDiff with representative state-of-the-art RRG methods, covering  
  both conventional encoder--decoder models and recent large multimodal models. For the multi-scenario MIMIC-RG4        
  benchmark, we follow the evaluation protocol of LLM-RG4~\citep{b5} and compare against the baselines reported or      
  reproduced under the same SN, SW, MN, and MW settings.

% Furthermore, given the rapid evolution of non-autoregressive text generation paradigms, it is necessary to clarify the distinctions between AnchorDiff and recently emerged concurrent diffusion-based models. Although pioneering works such as ECHO~\citep{b26} have begun exploring discrete diffusion mechanisms, their optimization objectives are primarily concentrated on accelerating parallel inference via uniform masking; while others, like DiagDiff~\citep{b27}, which attempt to incorporate priors, mostly rely on shallow feature concatenation. Moreover, these non-autoregressive methods are generally confined to the traditional SN task setting. In contrast, AnchorDiff is not only the first framework to introduce fine-grained clinical topological constraints (AnchorTree) and a dynamic inference-time correction mechanism (CAPTR) into the masked diffusion process, but it also pioneers the extension of the diffusion generation paradigm to complex clinical scenarios involving multiple views and historical records. Distinct from existing methods that prioritize decoding speed, our research focuses on safeguarding the clinical logic and anatomical structural accuracy of the generated reports.
\subsection{Implementation Details}
AnchorDiff is built upon LLaDA-8B-Instruct \citep{b21}, which serves as the masked diffusion language backbone. Visual features are extracted by a frozen RAD-DINO \citep{b22} encoder and subsequently projected into the LLaDA embedding space via a cross-attention projector with 128 learnable query tokens. Prior reports are independently encoded by a frozen CXR-BERT \citep{b23}. 

Model training follows a two-stage curriculum: Stage~1 aligns the multimodal projectors with the frozen backbone on single-view data by fine-tuning both the cross-attention modules and the MLPs; Stage~2 jointly fine-tunes all components across all four scenarios using LoRA \citep{b11}. Both stages employ the AdamW optimizer with an effective batch size of 32, where the learning rate is set to 1e-3 for Stage~1 and 3e-4 for Stage~2. The entire training pipeline can be efficiently completed on a single NVIDIA A800 GPU (80GB). AnchorTree is parameterized with $\beta{=}1.0$, $\gamma{=}1.5$, $\lambda{=}1.1$; CAPTR is configured with $M{=}3$, $E{=}8$, $\tau{=}0.3$, and is precisely activated within the 25\%-75\% core interval of the entire denoising process.

\section{Results and Analysis}
\subsection{Overall Results}

  Tables~\ref{tab:sota_sn} and~\ref{tab:multi_scenario} summarize the main results. In the standard SN setting,
  AnchorDiff achieves the best overall clinical efficacy among the compared methods. Compared with the strongest
  previous CheXbert F$_1$ baseline, AnchorDiff improves F$_1$ by approximately 3.6\%. This gain is accompanied by
  consistently strong NLG performance: on the cleaned-reference evaluation, AnchorDiff improves BLEU-1 and BLEU-4 over
  LLM-RG4 by about 4.8\% and 3.0\%, respectively. These results suggest that the proposed masked-diffusion formulation
  improves clinical correctness while preserving report-level language quality.

  The multi-scenario results further reveal where AnchorDiff is most effective. Across the four MIMIC-RG4 settings,
  AnchorDiff achieves an average relative precision gain of about 6.2\% over LLM-RG4 and an average BLEU-4 gain of about
  2.9\%. The improvement is especially clear in the no-prior settings. In SN and MN, where generation must rely mainly
  beneficial when the model cannot rely on historical textual context and must organize the report from visual evidence.

  In the history-aware settings, AnchorDiff remains competitive in clinical efficacy while maintaining consistent gains
  on most BLEU-based metrics. This indicates that the proposed framework can still benefit from additional textual
  context, but its relative advantage is more pronounced when image-grounded reasoning plays a larger role. We further
  analyze the precision--recall behavior behind this phenomenon in the following discussion.

  Overall, the results show that AnchorDiff does not simply improve one isolated metric. Instead, it produces a
  consistent performance pattern: stronger clinical specificity, improved BLEU-based generation quality, and larger
  gains in visually grounded no-prior scenarios. This pattern aligns with our design hypothesis that masked diffusion,
  when guided by clinical anchor structures, can reduce the dependence on fixed-order report templates and generate
  reports in a more image-conditioned manner.
 \subsection{Discussion on RRG Metrics and Sequence Bias}

  The results suggest that RRG performance should be interpreted through both clinical precision and recall. Recall
  rewards the coverage of reference-positive findings, whereas precision penalizes unsupported abnormalities. In
  history-aware settings, autoregressive models can benefit from prior reports by copying or paraphrasing previous
  findings, which may improve recall but can also reinforce report-level shortcuts. In contrast, AnchorDiff shows
  stronger precision-oriented behavior, indicating a more conservative generation process grounded in current evidence.

  This trend is consistent with our sequence-bias motivation. Autoregressive decoding conditions each token on the
  generated prefix, making later predictions prone to follow common report patterns once an early trajectory is formed.
  AnchorDiff mitigates this limitation by using bidirectional masked denoising and RadGraph-derived clinical anchors,
  which help organize generation around clinically meaningful structures rather than a fixed textual order. The stronger
  gains in no-prior scenarios further support this interpretation, since these settings rely more directly on current
  image evidence.

\begin{table*}[t]
\centering
% 使用 resizebox 将表格宽度强制缩放至页面文本总宽度
\resizebox{\textwidth}{!}{%
\begin{tabular}{l ccc ccc ccc}
\toprule
\multirow{2}{*}{\textbf{Model}} & \multicolumn{3}{c}{\textbf{CE Metrics}} & \multicolumn{3}{c}{\textbf{Clean NLG}} & \multicolumn{3}{c}{\textbf{Original NLG}} \\
\cmidrule(lr){2-4} \cmidrule(lr){5-7} \cmidrule(lr){8-10}
& P & R & F1 & B@1 & B@4 & R-L & B@1 & B@4 & R-L \\
\midrule
R2Gen \citep{b1} & 0.456 & 0.306 & 0.366 & 0.363 & 0.090 & 0.269 & 0.356 & 0.097 & 0.267 \\
R2GenCMN \citep{b2} & 0.486 & 0.400 & 0.439 & 0.385 & 0.102 & 0.278 & 0.349 & 0.094 & 0.270 \\
RGRG$^{\dagger}$ \citep{b31} & 0.461 & 0.475 & 0.447 & - & - & - & 0.373 & 0.126 & 0.264 \\
R2GenGPT \citep{b32} & 0.506 & 0.414 & 0.456 & 0.401 & 0.118 & 0.277 & 0.396 & 0.113 & 0.273 \\
EKAGen$^{\dagger}$ \citep{b33} & 0.517 & 0.483 & 0.499 & - & - & - & 0.419 & 0.117 & 0.287 \\
Promptmrg \citep{b34} & \colorbox{blue!10}{0.618} & 0.491 & 0.548 & 0.326 & 0.080 & 0.261 & 0.381 & 0.096 & 0.258 \\
CheXagent \citep{b35} & 0.506 & 0.306 & 0.381 & 0.265 & 0.058 & 0.239 & 0.189 & 0.040 & 0.208 \\
Med-PaLM$^{\dagger}$ \citep{b36} & - & - & 0.516 & - & - & - & 0.317 & 0.115 & 0.275 \\
UAR$^{\dagger}$ \citep{b41} & - & - & - & - & - & - & 0.363 & 0.107 & 0.289 \\
R2-LLM$^{\dagger}$ \citep{b37} & 0.465 & 0.482 & 0.473 & - & - & - & 0.402 & 0.128 & 0.291 \\
InVERGe$^{\dagger}$ \citep{b38} & - & - & - & - & - & - & 0.425 & 0.100 & 0.309 \\
MPO$^{\dagger}$ \citep{b39} & 0.436 & 0.376 & 0.353 & - & - & - & 0.416 & 0.139 & 0.309 \\
CPO \citep{b40} & - & - & - & - & - & - & \colorbox{blue!10}{0.426} & \colorbox{blue!10}{0.149} & \colorbox{blue!10}{0.321} \\
MARE$^{\dagger}$ \citep{b42} & 0.433 & 0.378 & 0.375 & - & - & - & 0.409 & 0.133 & 0.291 \\
RadSCR$^{\dagger}$ \citep{b44} & 0.440 & 0.433 & 0.408 & - & - & - & - & - & - \\
MLRG \citep{b43} & 0.549 & 0.468 & 0.505 & - & - & - & 0.411 & 0.147 & 0.320 \\
LLM-RG4 \citep{b5} & 0.583 & \colorbox{blue!10}{0.593} & \colorbox{blue!10}{0.588} & \colorbox{blue!10}{0.476} & \colorbox{blue!10}{0.197} & \textbf{0.385} & 0.377 & 0.144 & 0.318 \\
\midrule
\textbf{AnchorDiff(Ours)} & \textbf{0.621} & \textbf{0.599} & \textbf{0.609} & \textbf{0.499} & \textbf{0.203} & \colorbox{blue!10}{0.379} & \textbf{0.428} & \textbf{0.152} & \textbf{0.324}\\
\bottomrule
\end{tabular}%
}
 \caption{Comparison with SOTA methods for the setting of sn. $^{\dagger}$ indicates the results are quoted from the published literature. Clean NLG refers to using the cleaned reports from MIMIC-RG4 as ground truth, while Original NLG denotes using the original reports from MIMIC-
CXR as ground truth. The best results are in bold. The second-best results are highlighted in blue.}
\label{tab:sota_sn}
\vspace{-5pt}
\end{table*}

\begin{table*}[t]
\centering
% 使用 resizebox 确保双栏排版时不超宽
\resizebox{\textwidth}{!}{%
\begin{tabular}{c l ccc cccccc}
\toprule
\multirow{2}{*}{\textbf{Dataset}} & \multirow{2}{*}{\textbf{Model}} & \multicolumn{3}{c}{\textbf{CE Metrics}} & \multicolumn{6}{c}{\textbf{NLG Metrics}} \\
\cmidrule(lr){3-5} \cmidrule(lr){6-11}
& & P & R & F1 & B@1 & B@2 & B@3 & B@4 & R-L & MTR \\
\midrule

\multirow{4}{*}{\begin{tabular}[c]{@{}c@{}} \Large\textbf{\textit{sn}}\end{tabular}} 
& cxrmate$^*$ & 0.572 & 0.560  & 0.566  & 0.421  & 0.271  & 0.179  & 0.122  & 0.311  & 0.174  \\
& RadFM       & 0.413  & 0.303  & 0.350  & 0.188  & 0.090  & 0.048  & 0.028  & 0.190  & 0.094  \\
& LLM-RG4$^*$ & \colorbox{blue!10}{0.583} & \colorbox{blue!10}{0.593} & \colorbox{blue!10}{0.588} & \colorbox{blue!10}{0.476} & \colorbox{blue!10}{0.342} & \colorbox{blue!10}{0.255} & \colorbox{blue!10}{0.197} & \textbf{0.385} & \colorbox{blue!10}{0.207} \\
\cmidrule(lr){2-11}
& \textbf{Ours} & \textbf{0.621} & \textbf{0.599} & \textbf{0.609} & \textbf{0.499} & \textbf{0.360} & \textbf{0.267} & \textbf{0.203} & \colorbox{blue!10}{0.379} & \textbf{0.213} \\
\midrule

\multirow{4}{*}{\begin{tabular}[c]{@{}c@{}} \Large\textbf{\textit{sw}}\end{tabular}} 
& cxrmate$^*$ & 0.573  & 0.549  & 0.561  & 0.361  & 0.220  & 0.139  & 0.093  & 0.284  & 0.153  \\
& RadFM       & 0.508  & 0.365  & 0.425  & 0.211  & 0.103  & 0.056  & 0.033  & 0.183  & 0.105  \\
& LLM-RG4$^*$ & \colorbox{blue!10}{0.596} & \textbf{0.622} & \textbf{0.610} & \colorbox{blue!10}{0.453} & \colorbox{blue!10}{0.320} & \textbf{0.237} & \colorbox{blue!10}{0.184} & \textbf{0.382} & \colorbox{blue!10}{0.199} \\
\cmidrule(lr){2-11}
& \textbf{Ours} & \textbf{0.612} & \colorbox{blue!10}{0.603} & \colorbox{blue!10}{0.607} & \textbf{0.456} & \textbf{0.322} & \colorbox{blue!10}{0.234} & \textbf{0.189} & \colorbox{blue!10}{0.358} & \textbf{0.208} \\
\midrule

\multirow{4}{*}{\begin{tabular}[c]{@{}c@{}} \Large\textbf{\textit{mn}}\end{tabular}} 
& cxrmate$^*$ & \colorbox{blue!10}{0.544}  & 0.522  & 0.533  & 0.437  & 0.289  & 0.199  & 0.141  & 0.332  & 0.179  \\
& RadFM       & 0.323  & 0.187  & 0.237  & 0.246  & 0.113  & 0.060  & 0.034  & 0.194  & 0.104  \\
& LLM-RG4$^*$ & 0.543 & \textbf{0.567} & \colorbox{blue!10}{0.554} & \colorbox{blue!10}{0.487} & \colorbox{blue!10}{0.356} & \colorbox{blue!10}{0.272} & \colorbox{blue!10}{0.215} & \colorbox{blue!10}{0.400} & \colorbox{blue!10}{0.213} \\
\cmidrule(lr){2-11}
& \textbf{Ours} & \textbf{0.586} & \colorbox{blue!10}{0.537} & \textbf{0.561} & \textbf{0.505} & \textbf{0.374} & \textbf{0.285} & \textbf{0.224} & \textbf{0.402} & \textbf{0.221} \\
\midrule

\multirow{4}{*}{\begin{tabular}[c]{@{}c@{}} \Large\textbf{\textit{mw}}\end{tabular}} 
& cxrmate$^*$ & \colorbox{blue!10}{0.548}  & 0.499  & 0.523  & 0.379  & 0.241  & 0.158  & 0.110  & 0.305  & 0.159  \\
& RadFM       & 0.456  & 0.297  & 0.360  & 0.191  & 0.095  & 0.054  & 0.034  & 0.178  & 0.095  \\
& LLM-RG4$^*$ & 0.536 & \textbf{0.563} & \colorbox{blue!10}{0.549} & \colorbox{blue!10}{0.455} & \colorbox{blue!10}{0.326} & \textbf{0.245} & \colorbox{blue!10}{0.194} & \colorbox{blue!10}{0.398} & \textbf{0.201} \\
\cmidrule(lr){2-11}
& \textbf{Ours} & \textbf{0.578} & \colorbox{blue!10}{0.519} & \textbf{0.550} & \textbf{0.458} & \textbf{0.332} & \colorbox{blue!10}{0.241} & \textbf{0.197} & \textbf{0.401} & \colorbox{blue!10}{0.200} \\

\bottomrule
\end{tabular}%
}
\caption{Comparative analysis of SOTA baselines supporting MIMIC-RG4 in four configurations. $^*$ indicates the model is retrained on MIMIC-RG4. The best results are in bold. The second-best results are highlighted in blue.}
\label{tab:multi_scenario}
\end{table*}
\subsection{Ablation Analysis}            \label{sec:Ablation}                                                                                                  
  Table~\ref{xiaorong} reports the ablation results of the two core components in AnchorDiff: AnchorTree 
  and CAPTR. Starting from the plain masked-diffusion baseline, adding CAPTR alone leads to only minor changes,         
  improving F-14 from 0.527 to 0.529 and F-5 from 0.554 to 0.560, while slightly decreasing BLEU-4 from 0.117 to 0.114. 
  This indicates that inference-time rewriting by itself is not sufficient to substantially improve report generation;  
  its effectiveness depends on whether the model has already learned clinically meaningful structural priors during     
  training.                                                                                                             
We then analyze the two ablation variants of AnchorTree. The UM variant removes token-level masking protection while  
  retaining hierarchical loss weighting. Compared with the baseline, UM improves F-14 from 0.527 to 0.548 and BLEU-4    
  from 0.117 to 0.132, showing that assigning stronger supervision to clinically important tokens is beneficial. The L0-
  only variant further protects anatomical anchor tokens and improves F-14 to 0.564 and BLEU-4 to 0.147, suggesting that
  anatomical structures provide useful scaffolds for report generation. However, both variants remain clearly below the 
  full AnchorTree setting, indicating that radiology reports require a hierarchy over anatomical anchors, pathological  
  findings, and modifiers rather than isolated anatomical protection.                                                   
                                                                                                                        
  The full AnchorTree achieves substantially stronger performance, reaching 0.592 F-14 and 0.607 F-5 even without CAPTR.
  When combined with CAPTR, the full model obtains the best overall results, including 0.609 F-14, 0.611 F-5, 0.499     
  BLEU-1, and 0.203 BLEU-4. These results show that AnchorTree and CAPTR are complementary: AnchorTree provides         
  training-time clinical structure for the denoising process, while CAPTR further revises unstable token decisions      
  during inference.
  \begin{table}[t]
  \centering
  \setlength{\tabcolsep}{7pt}
  \begin{tabular}{l c cccc}
  \toprule
  \textbf{Training Strategy} & \textbf{CAPTR} & \textbf{F-14} & \textbf{F-5} & \textbf{B@1} & \textbf{B@4} \\
  \midrule
  Baseline & $\times$ & 0.527 & 0.554 & 0.401 & 0.117 \\
  Baseline & \checkmark & 0.529 & 0.560 & 0.412 & 0.114 \\
  UM & $\times$ & 0.548 & 0.566 & 0.423 & 0.132 \\
  UM & \checkmark & 0.570 & 0.604 & 0.459 & 0.169 \\
  L0-only & $\times$ & 0.564 & 0.582 & 0.449 & 0.147 \\
  L0-only & \checkmark & 0.578 & 0.593 & 0.476 & 0.172 \\
  Full AnchorTree & $\times$ & 0.592 & 0.607 & 0.487 & 0.193 \\
  \textbf{Full AnchorTree} & \checkmark & \textbf{0.609} & \textbf{0.611} & \textbf{0.499} & \textbf{0.203} \\
  \bottomrule
  \end{tabular}
  \vspace{10pt}
  \caption{Ablation study of AnchorTree and CAPTR. UM denotes the uniform-masking variant that removes token-level
  masking protection while retaining hierarchical loss weighting. L0-only denotes the variant that protects only
  anatomical anchor tokens without modeling the full clinical hierarchy.}
  \vspace{-15pt}
  \label{xiaorong}
  \end{table}
  
Table~\ref{tab:ablation_anchortree_canshu} studies the sensitivity of the AnchorTree masking hierarchy parameters $
  \beta$ and $\gamma$. The default configuration achieves the best overall performance across both clinical and NLG
  metrics. Deviating from this setting leads to clear degradation, especially in BLEU-4 and F1. This suggests that the
  masking hierarchy needs to balance two competing goals: preserving clinically important anchors as stable context and
  still forcing the model to reconstruct enough anchor-related tokens during training. Over- or under-protecting anchor
  tokens weakens this balance and reduces generation quality.
  
Table~\ref{tab:ablation_core_canshu} evaluates the hyperparameters of CAPTR. Compared with AnchorTree, CAPTR is less
  sensitive in clinical metrics, but its NLG performance can still be affected by overly aggressive rewriting settings.
  Increasing the candidate size or changing the activation interval produces moderate changes, while raising the
  confidence threshold reduces BLEU-4 more noticeably. The default configuration provides the best trade-off, suggesting
  that CAPTR works best as a conservative correction module that revises only a small number of contextually unstable
  tokens.
  
 \begin{table}[t]
 \centering

 \setlength{\tabcolsep}{8pt}
 \begin{tabular}{ccccccc}
 \toprule
 $\beta$ & $\gamma$ & \textbf{B@1} & \textbf{B@4} & \textbf{F1} & \textbf{P} & \textbf{R} \\
 \midrule
 0.5 & 0.5 & 0.457 & 0.129 & 0.587 & 0.593 & 0.584 \\
 \rowcolor{blue!10} \textbf{1.0} & \textbf{1.5} & \textbf{0.499} & \textbf{0.203} & \textbf{0.609} & \textbf{0.621} & \textbf{0.599}\\
 1.5 & 1.5 & 0.431 & 0.107 & 0.578 & 0.583 & 0.574\\

 1.0 & 1.0 & 0.493 & 0.189 & 0.597 & 0.605 & 0.590\\
 1.0 & 2.0 & 0.465 & 0.146 & 0.574 & 0.576 & 0.569\\
 \bottomrule
 \end{tabular}
 \vspace{10pt}
 \caption{Sensitivity analysis of AnchorTree hyperparameters on the SN benchmark.
 The default configuration is marked with highlighted in blue.}
 \label{tab:ablation_anchortree_canshu}
 \vspace{-10pt}
 \end{table}
 
\begin{table}[t]
 \centering
 \setlength{\tabcolsep}{7pt}
 \begin{tabular}{cccccccc}
 \toprule
 $M$ & $E$ & $\tau$ & \textbf{B@1} & \textbf{B@4} & \textbf{F1} & \textbf{P} & \textbf{R} \\
 \midrule
 1   & 8  & 0.3 & 0.495 & 0.192 & 0.598 & 0.617 & 0.584\\
\rowcolor{blue!10} \textbf{3}   & \textbf{8}  & \textbf{0.3} & \textbf{0.499} & \textbf{0.203} & \textbf{0.609} & \textbf{0.621} & \textbf{0.599}\\
 5   & 8  & 0.3 & 0.491 & 0.195 & 0.601 & 0.614 & 0.589 \\
 \midrule
 3   & 16 & 0.3 & 0.493 & 0.196 & 0.604 & 0.616 & 0.593\\
 3   & 8  & 0.5 & 0.482 & 0.183 & 0.597 & 0.608 & 0.586\\
 \bottomrule
 \end{tabular}
 \vspace{10pt}
 \caption{Sensitivity analysis of CAPTR hyperparameters on the SN benchmark.
 The default configuration is marked with highlighted in blue.}
 \label{tab:ablation_core_canshu}
 \vspace{-10pt}
 \end{table}

\begin{figure}[h]
  \centering
  \includegraphics[width=1.0\linewidth]{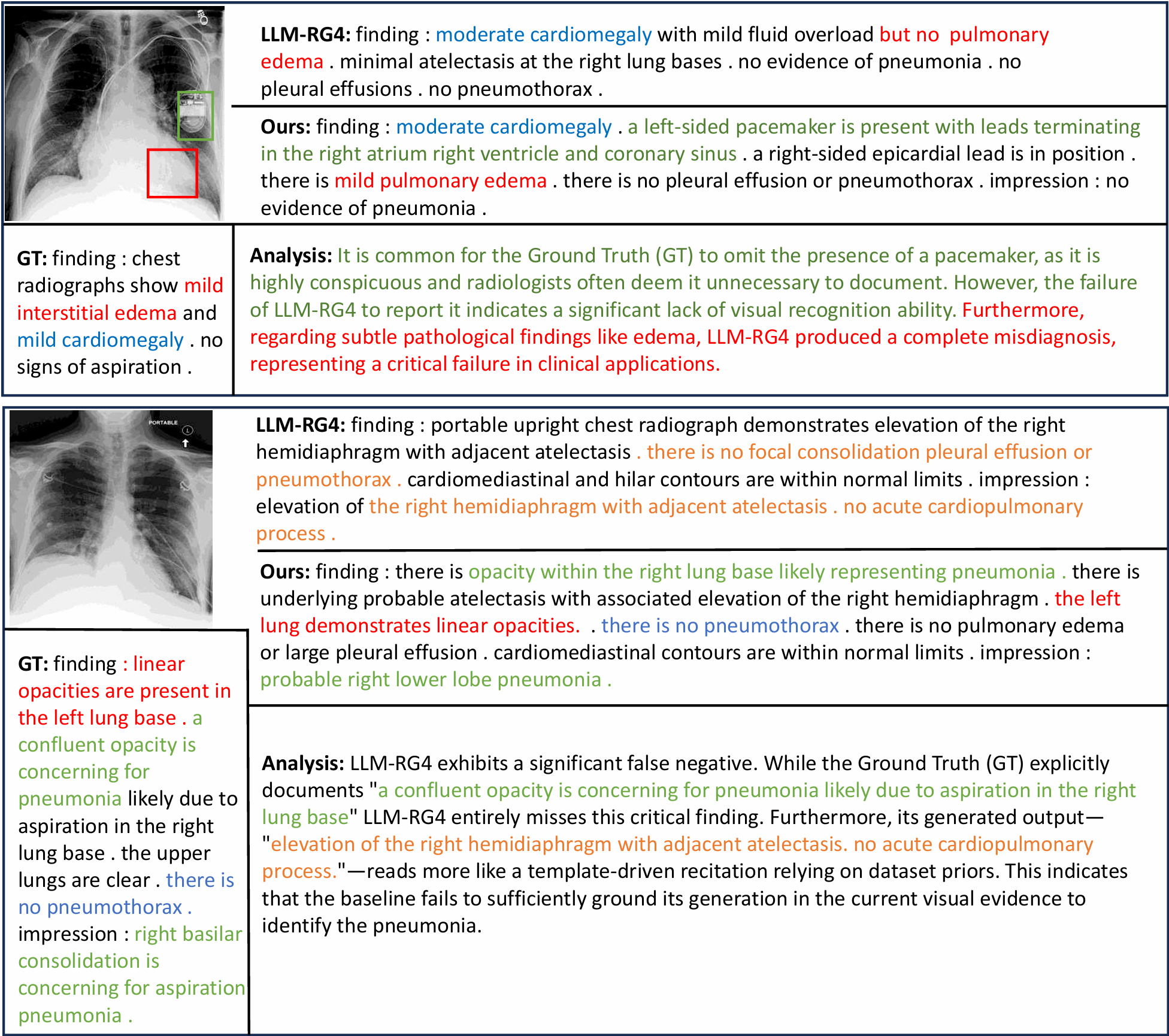}\href{}{}
  \caption{Qualitative case study. The upper example demonstrates AnchorDiff's ability to identify subtle abnormal      
  findings missed by LLM-RG4. The lower example illustrates the sequence bias of LLM-RG4, where a template-like report  
  contradicts the ground truth, while AnchorDiff produces a report that is more consistent with the image evidence.     
  Pathological findings correctly aligned with the ground truth are highlighted in matching colors, and missed or       
  contradicted findings are marked in yellow.}
  \label{fig:case_study}
\end{figure}
\begin{table*}[h]
\centering
% 使用 resizebox 防止表头过长导致表格超宽
\resizebox{\textwidth}{!}{%
\begin{tabular}{l cccccc}
\toprule
\textbf{Model} & \textbf{Lung Lesion} & \textbf{Edema} & \textbf{Atelectasis} & \textbf{Pneumothorax} & \textbf{Pleural Other} & \textbf{Fracture} \\
\midrule
Ours    & \textbf{0.741} & 0.559 & \textbf{0.770} & \textbf{0.780} & \textbf{0.620} & \textbf{0.868} \\
LLM-RG4 & 0.740 & \textbf{0.598} & 0.729 & 0.773 & 0.586 & 0.860 \\
\bottomrule
\end{tabular}%
}
\caption{Fine-grained CheXbert F1 comparison between AnchorDiff and LLM-RG4. AnchorDiff improves five out of six clinically important findings, especially localized or easily overlooked abnormalities such as atelectasis, pneumothorax, pleural other, and fracture.}
\label{tab:case_fine_grained}
\end{table*}
\subsection{Case Study}
To further examine whether the performance gains come from more faithful visual reasoning rather than superficial language matching, we provide both a fine-grained quantitative comparison and representative qualitative cases. As shown in Table~\ref{tab:case_fine_grained}, AnchorDiff achieves better F1 scores than LLM-RG4 on five out of six clinically important findings, including Lung Lesion, Atelectasis, Pneumothorax, Pleural Other, and Fracture. These categories are typically localized, subtle, or sparsely described, and are therefore more likely to be omitted when the model relies on dataset-level report priors. Although LLM-RG4 obtains a higher score on Edema, which often has more diffuse visual patterns and diverse textual expressions, AnchorDiff shows stronger overall sensitivity to fine-grained abnormalities that require explicit image grounding.                                                
          
Figure~\ref{fig:case_study} provides qualitative evidence for this behavior. In the upper example, AnchorDiff correctly identifies the subtle abnormal findings that are also described in the ground-truth report, while LLM-RG4 fails to mention them. This suggests that the proposed topology-aware masking strategy helps preserve clinically meaningful anchors and encourages the model to attend to local abnormal regions during generation.                    
                                                                                                                        
  The lower example further illustrates the sequence bias of autoregressive report generation. The ground-truth report  
  describes a clear abnormal finding, whereas LLM-RG4 produces a template-like report dominated by common normal-report 
  patterns and even states the absence of relevant abnormalities. In contrast, AnchorDiff generates a report that is    
  more consistent with the ground truth. This case supports our motivation that left-to-right autoregressive decoding   
  can overfit to high-probability report continuations, while the proposed diffusion-based formulation with confidence- 
  based rewriting better aligns report generation with image evidence.

\section{Conclusion} 
\label{sec:conclusion}

In this work, we presented \textbf{AnchorDiff}, a masked diffusion framework for radiology report generation designed to address the fundamental limitations of autoregressive models in clinical text synthesis. By replacing the conventional left-to-right generation paradigm with a bidirectional mask-and-predict process, AnchorDiff structurally eliminates sequence bias, a critical failure mode wherein models reproduce statistically frequent report patterns while neglecting patient-specific imaging evidence.

To fully exploit the non-sequential nature of masked diffusion, we introduced two complementary mechanisms that align the model's behavior with radiological reasoning. \textbf{AnchorTree} organizes the training objective around a RadGraph-derived clinical hierarchy, explicitly designating critical anatomical entities as anchors. By assigning topology-aware masking probabilities and loss weights, AnchorTree guides the model to treat these anchors as visible scaffolds, ensuring that pathological findings are accurately grounded in a robust anatomical framework. \textbf{CAPTR} operates at inference time to identify and revise \textit{context-brittle} tokens---those whose predicted identity shifts substantially under local context perturbation, providing an effective mechanism for post-hoc refinement. Together, these two mechanisms introduce minimal overhead: AnchorTree adds no inference cost, while CAPTR requires only a small number of additional forward passes at targeted decoding steps. Experiments on MIMIC-CXR demonstrate that AnchorDiff achieves competitive performance across standard evaluation metrics, validating the effectiveness of anchor-oriented, clinically grounded diffusion for structured medical report generation.

\bibliographystyle{plainnat} 
\bibliography{references.bib}

%%%%%%%%%%%%%%%%%%%%%%%%%%%%%%%%%%%%%%%%%%%%%%%%%%%%%%%%%%%%

\appendix

\end{document}